\documentclass[10pt,twocolumn,letterpaper]{article}

\usepackage{cvpr}
\usepackage{times}
\usepackage{epsfig}
\usepackage{graphicx}
\usepackage{amsmath}
\usepackage{amssymb}
\usepackage{csquotes}

\usepackage[breaklinks=true,bookmarks=false]{hyperref}

\cvprfinalcopy 

\def\cvprPaperID{****} 

\begin{document}

\title{Neonatal Pain Expression Recognition Using Transfer Learning}

\author{Ghada Zamzmi, Dmitry Goldgof, Rangachar Kasturi, and Yu Sun\\
Computer Science and Engineering, University of South Florida \\
Tampa, FL, 33620\\
{\tt\small \{ghadh,goldgof,r1k,yusun\}@mail.usf.edu}}

\maketitle

\begin{abstract}
Transfer learning using pre-trained Convolutional Neural Networks (CNNs) has been successfully applied to images for different classification tasks. In this paper, we propose a new pipeline for pain expression recognition in neonates using transfer learning. Specifically, we propose to exploit a pre-trained CNN that was originally trained on a relatively similar dataset for face recognition (i.e., VGG-Face) as well as CNNs that were pre-trained on a relatively different dataset for image classification (i.e., VGG-F,M,S) to extract deep features from neonates’ faces. In the final stage, several supervised machine learning classifiers are trained to classify neonates’ facial expression into pain or no pain expression. The proposed pipeline achieved, on a testing dataset, 0.841 AUC and 90.34\% accuracy, which is approx. 7\% higher than the accuracy of handcrafted traditional features. We also propose to combine deep features with traditional features and hypothesize that the mixed features would improve pain classification performance. Combining deep features with traditional features achieved 92.71\% accuracy and 0.948 AUC. These results show that transfer learning, which is a faster and more practical option than training CNN from the scratch, can be used to extract useful features for pain expression recognition in neonates. It also shows that combining deep features with traditional handcrafted features is a good practice to improve the performance of pain expression recognition and possibly the performance of similar applications.
\end{abstract}


\section{Introduction}
Infants receiving care in the Neonatal Intensive Care Unit (NICU) might experience up to several hundred painful procedures during their stay \cite{cruz2016epidemiology}. Pediatric studies have reported several long-term outcomes of repeated pain exposure in early life. For instance, it has been found \cite{grunau2006long} that repeated painful experience in neonates is associated with alterations in the cerebral white matter and subcortical grey matter and delayed cortico-spinal development. These alterations in neurodevelopment can result in a variety of behavioral, developmental and learning disabilities \cite{grunau2006long}. Other long-term outcomes of pain exposure that are reported \cite{field2017preterm} at school age include delayed visual–perceptual development, lower IQs, and internalizing behavior. 

The recognition of the adverse outcomes associated with neonatal pain exposure has led to the recommendation of using opioids such as Fentanyl and Morphine. Although analgesic medications can reduce the consequences of neonatal pain exposure, recent studies found a link between the excessive use of these medications and many short- and long-term side effects. Zwicker et al. \cite{zwicker2016smaller} found that 10-fold increase in Morphine, an agent commonly used for neonatal pain management, is associated with impaired cerebellar growth in the neonatal period and poorer neurodevelopmental outcomes in early childhood period. The long-term side effects of another well-known analgesic medication (i.e., Fentanyl) were discussed in \cite{tam2011preterm}. This study described Fentanyl as an extremely potent analgesic and listed several side effects, such as neuroexcitation, respiratory depression, for using high doses of Fentanyl.

These results suggest that the failure to recognize and treat pain when needed (i.e., under treatment) as well as the administration of analgesic medications in the absence of pain (i.e., over treatment) can cause serious outcomes and permanently changes the brain structure and functions. The annual cost of care related to adverse neurodevelopmental outcomes in preterm infants alone is estimated at over 7 billion dollars \cite{butler2007preterm}.

Because pain assessment is the cornerstone of pain management, the assessment of neonatal pain should be accurate and continuous. Currently, caregivers assess neonatal pain by observing behavioral (e.g., facial expression and crying) and physiological (e.g., vital signs changes) indicators using multidimensional pain scales such as NIPS (Neonatal Infant Pain Scale) \cite{hudson2002validation}, FLACC (Face, Legs, Activity, Crying, and Consolability) \cite{voepel2002reliability}, and NFCS (Neonatal Facial Coding System) \cite{peters2003neonatal}. This practice is inconsistent because it depends highly on the observer bias. Additionally, it is discontinuous and requires a large number of well-trained nurses to ensure proper utilization of the tools. The discontinuous nature of the current practice as well as the inter-rater variation may result in delayed intervention and inconsistent treatment of pain. Therefore, developing automated and continuous tools that can generate immediate and more consistent pain assessment is crucial.

\section{Existing Work and Contribution}
The recent innovations in computer vision facilitated the development of automated approaches that continuously and consistently assess pain. A large body of methods has been proposed to automatically assess pain using behavioral (e.g., facial expression \cite{liu2017deepfacelift,LITTLEWORT20091797,Hammal:2012:ADP:2388676.2388688,HAMMAL20121265,4042206,lucey2008improving,ASHRAF20091788,5771462,6553762,zhu2014pain,niese2009towards,LITTLEWORT20091797,BARTLETT2014738,sikka2015automated,werner2014comparative,florea2014learning,werner2013towards,martinez2017338,RATHEE201677,BRAHNAM2006211,BRAHNAM20071242,5415598,nanni2010local,zamzami2015pain} and crying \cite{Pal1660444,pai2016automatic}) or physiological (e.g., changes in vital signs \cite{faye2010newborn,gruss2015pain} and cerebral hemodynamic changes \cite{brown2011towards,RANGER2014519}) indicators. The vast majority of these methods assess and estimate pain based on analysis of facial expression. This focus is due to the fact that facial expression is the most common and specific indicator of pain \cite{hadjistavropoulos1997judging}. As such, most pediatric pain scales \cite{hudson2002validation, voepel2002reliability,peters2003neonatal} include facial expression as a main indicator for pain assessment.

Of the existing methods for automatic pain expression analysis, only few methods \cite{BRAHNAM20071242,nanni2010local,zamzami2015pain} focused on neonatal pain. This can be attributed to the lack of publicly-available neonatal datasets. Another reason is the common belief that the algorithms designed for adults should have similar performance when applied in neonates. Contrary to this belief, we think the methods designed for assessing adults’ pain will not have similar performance and might completely fail for two main reasons. First, the facial morphology and dynamics vary between infants and adults as reported \cite{oster2006baby}. Moreover, infants’ facial expressions include additional movements and units that are not present in the Facial Action Coding System. As such, Neonatal FACS was introduced as an extension of FACS \cite{oster2006baby,peters2003neonatal}. Second, we think the preprocessing stage (e.g., face tracking) is more challenging in the case of infants because they are uncooperative subjects recorded in an unconstrained environment.

The methods of automatic recognition of neonatal pain expression can be divided into two main categories: static and dynamic methods. 

Static methods extract pain-relevant features from static images and use the extracted features to train off-the-shelf
classifiers. One of the first work that detects and classify pain expression from infants' images (COPE dataset) is presented in \cite{BRAHNAM20071242}. The proposed method takes a static image as input and concatenates it into a feature vector of $ Image_{W} \times Image_{H}$ dimensions with values ranging from 0 to 255. Then, Principal Component Analysis (PCA) was applied to reduce the vector's dimensionality. For classification, distance-based classifiers and Support Vector Machines (SVMs) were used to classify the images into one of the following four pair: pain/no-pain, pain/cry, pain/air puff, and pain/friction. The results showed that SVMs evaluated using 10-fold cross-validation achieved the best recognition rate and outperformed distance-based classifiers in classifying pain versus no-pain (88.00\%), pain versus rest (94.62\%), pain versus cry (80.00\%), pain versus air-puff (83.33\%), and pain versus friction (93.00\%). This work was extended \cite{Brahnam2007} by employing Sequential Floating Forward Selection for feature selection and Neutral Network Simultaneous Optimization Algorithm (NNSOA) for classification, and an average classification rate of 90.2\% was obtained. Nanni et al. \cite{nanni2010local} applied several variations of Local Binary Pattern (LBP) on static images of the COPE dataset to classify them into pain and no-pain expression. These variations include Local Ternary Pattern (LTP), Elongated Local Ternary Pattern (ELTP), and Elongated Local Binary Pattern (ELBP). The highest performance was achieved by ELTP with AUC (Area under the Curve of Receiver Operating Characteristic Curve) score of 0.93. A complete review of the exiting methods for pain expression recognition can be found in \cite{zamzmi2016machine}.

The above-listed works utilize traditional handcrafted features for classification. Recently, deep feature extracted from a Convolutional Neural Networks (CNN) showed good performance in several classification tasks. The main difference between handcrafted features and deep features is that the features extracted by CNN are learned, at multiple levels of abstraction, directly from the data in contrast to the handcrafted features that are designed beforehand by human experts to extract a given set of chosen characteristics. 

This paper contributes a novel pipeline to recognize pain expression in neonates using transfer learning. Specifically, we propose to use four pre-trained Convolutional Neural Networks (CNNs) architectures, namely VGG-F, VGG-M, VGG-S, and VGG-Face, and show that these pre-trained CNNs can be used to extract useful features for pain expression classification in neonates. VGG-F,M,S architectures were originally trained on ImageNet dataset (approx. 1.2M images and 1000 class) for image classification while VGG-Face was trained on a large Face dataset (approx. 2.6M face images of 2622 identities) for face recognition. We hypothesize that the architectures that were originally trained on ImageNet for image classification can be used to extract useful features for pain classification. We also hypothesize that VGG-Face can be used for pain classification and it would provide better performance than the first three architectures because it was pre-trained on a relatively similar dataset. The reason for choosing an architecture trained to recognize faces instead of emotions is that face recognition is well-studied and validated on large volume datasets as compared to emotion classification. Moreover, the features of face recognition and facial expression recognition are rather similar since both tasks involve analyzing human faces \cite{alexandr2017group}. 

In addition, this paper proposes a new approach to pain-emotion analysis that incorporates both deep features and traditional handcrafted features. We hypothesize that the mixed features can improve pain classification performance. 


\textbf{Organization:} Section 3 describes the infants' pain dataset utilized in this work. Section 4 presents the prepossessing stage of our proposed pipeline, provides brief introduction for Convolutional Neural Networks, and discusses how we used transfer learning for pain classification. Section 5 presents the experiments we designed to evaluate our hypotheses and summarizes the results. We conclude in section 6. 


\section{Infants Pain Assessment Database}
\subsection{Subject}
Infants (N = 31 infants, 16 female and 15\% male) were recorded undergoing a brief acute stimulus such as heel lancing or immunization during their hospitalization in the NICU at Tampa General Hospital. Infants' average gestational age is 36.4, ranging from 30.4 to 40.6 (SD = 2.7). The ethnic distribution was 17\% Caucasian, 47\% White, 17\% African American, 12\% Asian, and 7\% other. Any infant born in the range of 28 and 41 gestation weeks was eligible for enrollment after obtaining informed consent from the parents. We excluded infants with cranial facial abnormalities and neuromuscular disorders.

\subsection{Video Recording and Ground Truth Labeling}
We used GoPro Hero3+ camera to record infants' facial expression, body movement, and crying sound. All the recordings were carried out in the normal clinical environment that is only modified by the addition of the camera. 

We recorded each infant in seven time periods: 1) Prior the painful procedure to get the baseline state observation; 2) Procedure preparation period that begins with first touch, may include positioning or skin preparation and ends with skin breaking; 3) Painful procedure, lasts the duration of the procedure; 4) One minute post the completion of the painful procedure; 5) Two minute post the completion; 6) Three minute post the completion; and 7) Recovery period five minutes post the procedure. Each time period was observed by trained nurses to provide the pain assessment using NIPS (Neonatal Infant Pain Scale). 

NIPS scale consists of six elements, which are facial expression, crying, body movement (i.e., arms and legs), state of arousal, and breathing patterns. Each element of the NIPS was scored on a scale of 0-1 with the exception of cry, which is scored on a scale of 0-1-2. A total score of 3-4 represents moderate pain and a score $>$ 4 indicates severe pain. To get the ground truth for each video epoch, we used the thresholding of the total score (i.e., severe pain, moderate pain, or no pain) as the label for algorithm evaluation. In this paper, we included pain/no-pain labels and excluded moderate pain because the number of epochs for moderate pain in the current dataset is small. 



\section{Automatic Pain Expression Recognition}
The proposed pain expression recognition pipeline consists of three main stages: 1) face detection and preprocessing, 2) deep feature extraction using transfer learning, 3) feature selections and classification. We describe each stage in detail below. 

\subsection{Automatic Face Detection and Preprocessing}
We applied ZFace \cite{jeni2015dense}, which is a person-independent tracker, in each video to detect the face and obtain 49 facial landmark points. The tracker outputs the 49 points' coordinates as well as a failure message to indicate the failure frames; those frames were excluded from further analysis. For each frame, we used the detect points to register and crop the infant's exact face region. We applied the tracker in 200 videos to detect the face and the landmark points. Then, we selected the key frames from each video, thereby removed many similar frames. The selected frames from all videos (i.e., 3026 frames) were then resized to 224 X 224 to accommodate with CNNs image's size requirement (244 x 224 x 3, RGB images). 

\subsection{Deep Features Extraction}
In this section, we give a brief introduction to Convolutional Neural Networks (CNNs) and Transfer Learning. We also describe the pipeline we propose to extract useful features for pain classification. 

\subsubsection{Convolutional Neural Networks}
Convolutional Neural Networks (CNNs) gained a lot of popularity in the last decades due to the wide range of its successful applications in natural language processing, recommender systems, medical image analysis, object recognition, and emotion analysis. The power of CNNs, which are biologically-inspired variants of a multilayer perceptrons, can be attributed to its deep architecture that allows to extract a set of features (i.e., features independent of prior knowledge or human effort) at multiple levels of abstraction.

CNN consists of an input layer, an output layer, and three types of hidden layers:  convolutional layer, pooling layer, and fully connected layer. The convolutional layer applies k convolutional kernels or filters to the input and pass the result (i.e., feature map) to the next layer. This layer takes as input an $m \times m \times r$ image where m represents the image's width and height and r represent the number of channels (e.g., 3 channels for RGB); the filter's size of this layer is $n \times n$ where n $<$ m. The pooling layer takes each feature map as input and performs subsampling by taking the average or the maximum to create new subsampled feature map. The last type of hidden layers is the fully connected layer, which is a regular feed-forward Neural Network layer that computes the activation of each class; this layer is responsible for the high-level reasoning in the network.

In practice, it is more common to use a pre-trained CNN as a fixed feature extractor or as starting point (i.e., fine-tune the weights of pretrained CNNs) instead of training the network from the scratch due to two main reasons. First, it is relatively rare to find a labeled dataset that is large enough (e.g., ImageNet – approx. 1.2 million images and 1000 classes) to train CNNs from the scratch. The vast majority of the existing datasets, especially in the medical domain for neonatal population, are scarce. In fact, we are not aware of any publicly-available neonatal dataset, except the small COPE \cite{BRAHNAM2006211} dataset (204 face images), collected for pain assessment or similar application. Second, training CNNs requires an extensive computational and memory resources as well as patience and expertise to ensure the proper choice of architecture and learning parameters. Transfer learning is an alternative to training CNN from the scratch that has received much attention in machine learning research and practice \cite{paul2016deep,ng2015deep,pan2010survey}. Andrew Ng\footnote{\enquote{Transfer learning will be - after supervised learning - the next driver of ML commercial success}} described transfer learning as the next driver of machine learning commercial success. 

\begin{table}
\begin{center}
\begin{tabular}{|c|c|}
\hline
Conv 1 & $64 \times 11 \times 11$, st. 4, pad 0  \\
\hline
Conv 2 & $256 \times 5 \times 5$, st. 1, pad 2 \\
\hline
Conv 3 & $256 \times 3 \times 3$, st. 1, pad 1\\
\hline
Conv 4 & $256 \times 3 \times 3$, st. 1, pad 1\\
\hline
Conv 5 & $256 \times 3 \times 3$, st. 1, pad 1\\
\hline
Full 6 & 4096 dropout \\
\hline
Full 7 & 4096 dropout\\
\hline
Full 8 & 1000 softmax\\
\hline
\end{tabular}
\end{center}
\caption{VGG-F architecture \cite{chatfield2014return}; k x n x n indicates number of filters and their size; st. and pad indicate the convolution stride and the padding. Each layer, except Full 8, is followed by RelU.}
\end{table}

\begin{table}
\begin{center}
\begin{tabular}{|c|c|}
\hline
Conv 1 & $96 \times 7 \times 7$, st. 2, pad 0  \\
\hline
Conv 2 & $256 \times 5 \times 5$, st. 2, pad 1 \\
\hline
Conv 3 & $512 \times 3 \times 3$, st. 1, pad 1\\
\hline
Conv 4 & $512 \times 3 \times 3$, st. 1, pad 1\\
\hline
Conv 5 & $512 \times 3 \times 3$, st. 1, pad 1\\
\hline
Full 6 & 4096 dropout \\
\hline
Full 7 & 4096 dropout\\
\hline
Full 8 & 1000 softmax\\
\hline
\end{tabular}
\end{center}
\caption{VGG-M architecture \cite{chatfield2014return}; k x n x n indicates number of filters and their size; st. and pad indicate the convolution stride and the padding. Each layer, except Full 8, is followed by RelU.}
\end{table}

\begin{table}
\begin{center}
\begin{tabular}{|c|c|}
\hline
Conv 1 & $96 \times 7 \times 7$, st. 2, pad 0  \\
\hline
Conv 2 & $256 \times 5 \times 5$, st. 1, pad 1 \\
\hline
Conv 3 & $512 \times 3 \times 3$, st. 1, pad 1\\
\hline
Conv 4 & $512 \times 3 \times 3$, st. 1, pad 1\\
\hline
Conv 5 & $512 \times 3 \times 3$, st. 1, pad 1\\
\hline
Full 6 & 4096 dropout \\
\hline
Full 7 & 4096 dropout\\
\hline
Full 8 & 1000 softmax\\
\hline
\end{tabular}
\end{center}
\caption{VGG-S architecture \cite{chatfield2014return}; k x n x n indicates number of filters and their size; st. and pad indicate the convolution stride and the padding. Each layer,except Full 8, is followed by RelU.}
\end{table}

\subsubsection{Transfer Learning}
Transfer learning is about applying knowledge that is learned from a previous domain/task to a new relevant domain/task. It offers an attractive solution for the lack of large and annotated datasets issue, which is known to be common in medical application. The idea of transfer learning is inspired from human learning and the fact that people can intelligently learn or solve a new problem using previously learned knowledge. 

There are two main scenarios for transfer learning: fine-tuning and fixed feature extractor. The first scenario involves fine-tuning the weights of the higher layers in the pre-trained CNN by continuing backpropagation since these layers contain dataset-specific features while the lower layers contains generic features (e.g., edge detector and color). In the second scenario, the pre-trained CNN is used as a fixed feature extractor to extract deep features after removing the output layer. The extracted features will then be used to train supervised machine learning classifiers (e.g., SVM) for a new task.

In this paper, we propose a pipeline for neonatal pain expression recognition using the second scenario of transfer learning. Specifically, we used four pre-trained CNNs to extract deep features from our relatively small dataset (31 subjects, 3026 images). The first three CNNs architectures, which are VGG-F, VGG-M, and VGG-S \cite{chatfield2014return}, were previously trained on a subset of ImageNet dataset (approx. 1.2M images and 1000 classes) for image classification. Tables 1-3 provide the architectures for VGG-F, VGG-M, and VGG-S respectively. The fourth CNN architecture (depicted in Table 4) is VGG-Face descriptor \cite{parkhi2015deep}, which was previously trained on large face dataset (approx. 2.6M face images of 2622 identities) for face recognition. Choosing these pre-trained CNNs allows us to investigate the difference between using CNNs trained on a relatively similar dataset (i.e., VGG-Face, Face dataset) and CNNs trained on a relatively different dataset (i.e., VGG-F,M,S, ImageNet). We hypothesize that these pre-trained CNN architectures can be used to extract useful texture features for pain classification. 

For all the four pre-trained architectures, we extracted deep features from the last fully connected layer before the output layer (Full 7 in Tables 1-4) which has high-level features more relevant to the utilized dataset. We also extracted features from the last convolutional layer (Conv 5 in Tables 1-4) which has low and generic features.

\begin{table}
\begin{center}
\begin{tabular}{|c|c|}
\hline
Conv 1-1 & $64 \times 3 \times 3$, st. 1, pad 1  \\
Conv 1-2 & $64 \times 3 \times 3$, st. 1, pad 1 \\
\hline
Conv 2-1 & $128 \times 3 \times 3$, st. 1, pad 1  \\
Conv 2-2 & $128 \times 3 \times 3$, st. 1, pad 1 \\
\hline
Conv 3-1 & $256 \times 3 \times 3$, st. 1, pad 1  \\
Conv 3-2 & $256 \times 3 \times 3$, st. 1, pad 1 \\
Conv 3-3 & $256 \times 3 \times 3$, st. 1, pad 1  \\
\hline
Conv 4-1 & $512 \times 3 \times 3$, st. 1, pad 1 \\
Conv 4-2 & $512 \times 3 \times 3$, st. 1, pad 1  \\
Conv 4-3 & $512 \times 3 \times 3$, st. 1, pad 1  \\
\hline
Conv 5-1 & $512 \times 3 \times 3$, st. 1, pad 1 \\
Conv 5-2 & $512 \times 3 \times 3$, st. 1, pad 1  \\
Conv 5-3 & $512 \times 3 \times 3$, st. 1, pad 1  \\
\hline
Full 6 & 4096 dropout \\
\hline
Full 7 & 4096 dropout\\
\hline
Full 8 & 2622 \\
\hline
\end{tabular}
\end{center}
\caption{VGG-Face architecture \cite{parkhi2015deep}, st. and pad indicate the convolution stride and the padding. Each layer (e.g., Conv 1-1) followed by ReLU and each block (e.g., Conv 1-1 and Conv 1-2) followed by pooling. }
\end{table}

\subsection{Feature Selection and Classification}
The deep feature vector extracted using transfer learning is high in dimensions, and hence performing feature selection was necessary. In this section, we briefly present two feature selection methods as well as four machine learning classifiers.

\subsubsection{Feature Selection}
Feature selection methods aim to select, from a given feature vector, the most relevant features while discarding irrelevant or redundant features. In this paper, we used Relief-f and Symmetric Uncertainty methods. 

Relief-f \cite{kira1992practical} method searches for the neighbor from the same class (i.e., nearest hit) and a neighbor from the opposite class (i.e., nearest miss) for each instance using a nearest neighbor algorithm. It then selects features according to their weight, which increases or decreases as a function of how well the feature distinguishes between distinct classes. In our experiments, we used the best 5, 10, and 15 features for classification. 

Symmetric uncertainty is a feature selection method that measures feature-correlation. It selects features based on the hypothesis that, "Good features subsets contain features highly correlated with the class, uncorrelated to each other" \cite{hall1999correlation}. It is computed as follows \cite{hall1999correlation}: 

\begin{equation}
   SU = 2.0 \times \frac{H(X) + H(Y)-(X,Y)}{H(Y)+H(X)} 
\end{equation}

Where X and Y represent two features and H(X) and H(Y) represent the entropy of these features. This symmetric measure ranges from 0 (uncorrelated) to 1 (correlated). We used the best 5, 10, and 15 features found by the algorithm for classification. 

\subsubsection{Classification}
There exist a wide range of classification algorithms, each has its strengths and weakness. In this work, we experimented with Naive Bayes, Nearest Neighbors (kNN), Support Vector Machines (SVMs), and Random Forests (RF) classifiers because they have shown good classification performance in transfer learning applications \cite{paul2016deep,rassadin2017group,sargano2017human,bar2015deep,bar2015deep}. A brief overview of these classifiers is presented below. 

Naive Bayes is a simple yet efficient probabilistic classifier that depends on Bayes' Theorem. It simplifies learning because it does not require iterative parameter estimation and makes an assumption, given the class variable, that features are independent. The learning phase of this classifier involves estimating the conditional and prior probabilities for each class. To classify a new instance, Naive Bayes, given the feature values of this instance, computes the posterior probability for each class and assigned the given instance to the class that has the highest probability.

\begin{table*}
\begin{center}
\begin{tabular}{|l|c|c|c|c|c|c|c|c|}
\hline
CNNs Name & \multicolumn{2}{|c|}{VGG-F} & \multicolumn{2}{|c|}{VGG-M} & \multicolumn{2}{|c|}{VGG-S} & \multicolumn{2}{|c|}{VGG-Face} \\
\hline
& PostReLU & PreReLU & PostReLU & PreReLU & PostReLU & PreReLU & PostReLU & PreReLU \\
\hline
Dimensions & 4096 & 4096 & 4096 & 4096 & 4096 & 4096 & 4096 & 4096\\
\hline
Selection & RF(5) & SU(5) & SU(10) & SU(10) & RF(10) & SU(5) & SU(15) & SU(5) \\
\hline
Classifier & SVMs & NB & RFT & NB & NB & RFT & kNN & kNN\\
\hline
Accuracy & 83.86 & 89.29 & 83.13 & 83.86 & \textbf{90.41} & 87.10 & \textbf{90.34} & 89.55 \\
\hline
AUC & 0.741 & 0.744 & 0.740 & 0.750 &  \textbf{0.742} & 0.716 & \textbf{0.841} & 0.859 \\
\hline
\end{tabular}
\end{center}
\caption{Pain classification performance using deep features of higher layer\textsuperscript{\ref{2}}}
\end{table*}

\begin{table*}
\begin{center}
\begin{tabular}{|l|c|c|c|c|c|c|c|c|}
\hline
CNNs Name & \multicolumn{2}{|c|}{VGG-F} & \multicolumn{2}{|c|}{VGG-M} & \multicolumn{2}{|c|}{VGG-S} & \multicolumn{2}{|c|}{VGG-Face} \\
\hline
& PostReLU & PreReLU & PostReLU & PreReLU & PostReLU & PreReLU & PostReLU & PreReLU \\
\hline
Dimensions & 43264 & 43264 & 86528 & 86528 & 147968 & 147968 & 100352 & 100352\\
\hline
Selection & SU(10) & SU(15) & SU(10) & SU(15) & RF(15) & SU(10) & SU(15) & SU(10) \\
\hline
Classifier & NB & NB & NB & RFT & NB & RFT & NB & kNN\\
\hline
Accuracy & \textbf{87.13} & 84.72 & 86.32 & 83.06 & 86.31 & 84.13 & \textbf{88.23} & 82.47 \\
\hline
AUC &  \textbf{0.713} &  0.764 & 0.754 & 0.663 & 0.711 & 0.703 & \textbf{0.797} & 0.700 \\
\hline
\end{tabular}
\end{center}
\caption{Pain classification performance using deep features of lower layer\textsuperscript{\ref{2}}}
\end{table*}

Nearest neighbor is a non-parametric machine learning algorithm that classifies a new instance based on the class of its nearest instances (i.e., neighbors). The classification phase for kNN is delayed to run-time, hence it is also known as a lazy classifier. To find the neighbors for a new instance, a distance metric (e.g., Euclidean distance) is computed between the given instance and k neighbor instances. Then, a majority voting is applied on the k neighbor to choose the most common class as the class for the new instance. This algorithm is simple and effective, but requires large memory space because it needs all the instances to be in memory at run-time. 

SVMs is a supervised classifier that performs classification by finding the optimal separating hyperplane that maximizes the margin or the distance between two classes' closest points (i.e., support vectors); removing those points would change the position of the hyperplane. The mathematical formulation of SVMs and more discussion about it can be found in \cite{hearst1998support}.

Random forest is a supervised classification algorithm that constructs, at training time, ensembles of decision trees (i.e., forest of trees) and chooses the mode class among all the trained trees. It uses bootstrapping on the training set and random feature selection in the tree induction. This method can run efficiently on large datasets with thousands of features.


\section{Experiments and Results}
To classify infants' facial expression as pain or no pain expression, a total of 3026 face images were fed to the four previously-mentioned CNNs architectures to extract deep features. All CNNs are implemented in a MATLAB Toolbox known as MatConvNet \cite{vedaldi2015matconvnet}.

The extracted deep features were then divided into training (16 subjects, 1514 images/instances) and testing (1512, 15 subjects) sets. For feature selection and classification, we applied feature selectors on the training set followed by machine learning classifiers. We experimented with the feature selection methods and the classifiers as implemented in Weka (version 3.7.13). 

We divide the experiments into three main folds: deep features from higher-layer, deep features from lower-layer, and merging deep features with traditional features. The reason for extracting features from both higher and lower layers is to investigate "what is the best layer to transfer?". Then, we combined the deep features extracted using transfer learning with traditional features extracted as described in \cite{zamzmi2017automated}. We present each experiment and report its results next.

\subsection{Higher Layer Deep Features}
Higher layers (i.e., closer to the output) in CNNs contain high level features that are specific to the utilized dataset. We hypothesize that the deep features extracted from higher fully connected layer of CNNs trained for image classification (i.e., VGG-F,M,S) can be used for pain classification. In our experiment, we removed the output softmax layer (i.e., Full 8 in Tables 1-3) and extracted deep features from Full 7 before applying Rectified Linear Unit function (PreReLU features, 4096 dimensions). We also extracted the features after they have been transformed by a ReLU function (PostReLU features, 4096 dimensions). The results of classifying pain using these features are summarized in the first three columns of Table 5\footnote{NB, RFT, RF, and SU represent Naive Byes, Random Forest Trees, RelieF, and Symmetric Uncertainty; (\#) indicate number of selected features. \label{2}}. As can be seen from the table, the highest accuracy for pain classification (90.41\%) was obtained with the PostReLU features extracted from VGG-S architecture. Although PostReLU features of VGG-S has the highest accuracy, assessing the significance of the difference between AUC of VGG-S (0.742) and AUCs of VGG-F,M showed no significant difference (P=0.05).

In addition to VGG-F,M,S, we extracted deep features from the last fully connected layer of VGG-face CNNs (i.e., Full 7 in Table 4) after removing the output layer. We extracted the features before and after applying ReLU (PreReLU features with 4096 dimensions and PostReLU with 4096 dimensions). We hypothesize that the features extracted using this architecture should achieve better results than VGG-F,M,S since it is trained originally on a dataset relatively similar to our infant's faces dataset. The last column of Table 5 presents the pain classification results using the deep features of VGG-face CNNs. The performance of pain classification using deep features extracted from VGG-Face achieved 90.34\% accuracy and 0.841 AUC. The AUC difference between VGG-Face (0.841) and VGG-S (0.742) is statistically significant at the P=0.05 level; the gray cells in Table 5 indicates this significant difference. This result is consistent with our hypothesis that VGG-Face would achieve better pain classification results. 

\subsection{Lower Layer Deep Features}
Contrary to the higher layers that have features customized according to the dataset used to train the CNN, lower layers contain generic low-level features (e.g., edge detector and colors) that are less specific to the utilized dataset. Experimenting with lower layers' features allow us to explore the usefulness of generic features for pain classification and compare their results with higher layers. 

In case of VGG-F,M,S, the deep features were extracted from the last convolutional layer (i.e., Conv 5 in Tables 1-3) before and after ReLU (PreReLU features and PostReLU features). The feature vectors dimensions are 43264, 86528, and 147968 for VGG-F, VGG-M, and VGG-S respectively. See the first three columns in Table 6 for a summary of performance. As can be seen from the table, the highest accuracy (87.13\%) was obtained with the PostReLU features extracted from VGG-F architecture. Although PostReLU features of VGG-F has the highest accuracy, assessing the significance of the difference between AUC of VGG-F (0.713) and AUCs of VGG-M,S showed no statistical difference (P=0.05).

We also extracted deep features from the last convolutional layer (Conv 5 in Table 4) of VGG-Face. The dimensions of the extracted feature vectors before and after applying ReLU is 100352 (see last column in Table 6). Using the lower-layer of VGG-Face for pain classification achieved 88.23\% accuracy and 0.797 AUC. The AUC difference between VGG-Face (0.797) and VGG-F (0.713) is statistically significant at the P=0.05 level as indicated by the gray cells in Table 6. 

As we mentioned earlier, the reason for extracting features from higher and lower layers is to investigate which layers would give better pain classification results. Therefore, we compared the best result obtained from higher-layer of VGG-F,M,S (i.e., VGG-S, 90,41 accuracy and 0.742 AUC) with the best result obtained from lower-layer of these three CNNs (i.e., VGG-F, 87.13 accuracy and 0.713 AUC). The higher-layer accuracy is approx. 3.7\% higher than the lower-layer. However, the AUC difference between them is not statistically significant at the P=0.05 level. Similarly, we compared the best result obtained from higher-layer of VGG-Face with the best result obtained from lower-layer. The former's accuracy is approx. 2.3\% higher than the latter, but the AUC difference between them is not statistically significant at the P=0.05 level.

\begin{table}
\begin{center}
\begin{tabular}{|c|c|c|c|}
\hline
Name & Strain & VGG-Face & Strain $+$ VGG-Face\\
\hline
Features \# & 5 & SU(15) & Strain(5)+Deep(15)  \\
\hline
Classifier & SVM & kNN & NB \\
\hline
Accuracy & 83.88 & 90.34 & \textbf{92.71} \\
\hline
AUC & 0.719 & 0.841 &  \textbf{0.948} \\
\hline
\end{tabular}
\end{center}
\caption{Pain classification performance using Mixed Features\textsuperscript{\ref{2}}}
\end{table}

\subsection{Merging Deep and Traditional Features}
In this experiment, we combined the top deep features of VGG-Face CNN architecture with traditional handcrafted features extracted using an optical-flow based method presented in \cite{zamzmi2017automated}. 

The optical flow based method works as follows. First, it calculates optical flow between consecutive frames of a video for the entire face region as well as for four regions (i.e., two upper regions and two lower regions). Then, it estimates the optical strain over the flow fields to generate the strain tensor components. Next, the strain magnitude is calculated for each region of the face along with the overall face region; each region generates a sequence (strain plot) corresponding to the amount of strain observed over time. Finally, the points of maximum strain are detected using a peak detector and the descriptive statistics for those peaks are calculated to generate the features (e.g., $FaceAll_{mean}$, $FaceI_{mean}$, $FaceII_{mean}$, $FaceIII_{mean}$, and $FaceIV_{mean}$). Using the strain features for pain classification gave 83.88\% accuracy and 0.719 AUC. 

Merging the deep features with the traditional strain features improve the pain classification performance. The best result (see Table 7, column 3) was obtained using a combination of five strain features and 10 PostReLU features extracted from the higher fully connected layer. This combination (Table 7, 4th column) showed $>$ 9\% increase in accuracy as compared to the accuracy of strain features (Table 7, 2nd column) and a statistically significant AUC difference (P=0.05). \newline 

To summarize, we present in this section three proposed experiments for neonatal pain classification using transfer learning. In the first two experiments, we extracted deep features from higher layer and lower layer of four pre-trained CNNs architectures. The higher layer features showed higher pain classification accuracy, but the AUC difference was not statistically significant at the P=0.05 level. The best pain classification results were obtained using VGG-Face architecture. This result is consistent with our hypothesis that VGG-Face would achieve better results than VGG-F,M,S since it was trained originally on a relatively similar dataset. In the last experiment, we combined deep features with traditional handcrafted extracted as described in \cite{zamzmi2017automated}. Using mixed features for pain classification yielded the best result with 92.71\% accuracy and 0.948\% AUC. 

We conclude, based on these preliminary results, that transfer learning can be used to extract useful features for pain classification in neonates. We also conclude that combining both traditional and deep features is a good practice to improve the performance of pain expression classification and possibly the performance of similar tasks. Though, further investigation, on a larger dataset, is required to validate these findings.

\section{Conclusion and Future Work}
This paper proposes a novel pipeline for neonatal pain expression recognition using pre-trained CNNs as feature extractor. The extracted feature vectors were used to train several machine learning classifiers after applying feature selections to select the most relevant features. The best result (90.34\% accuracy and 0.841 AUC) for pain expression recognition was obtained using deep features extracted from the last fully connected layer (Post-ReLU) after removing the output layer. This result is significantly higher (p=0.05) than the pain expression recognition using traditional handcrafted features (83.88\% accuracy and 0.719 AUC). Combining both handcrafted and deep features yielded 92.71\% accuracy and 0.948 AUC. These results conclude that transfer learning, which is a faster and more practical option than training CNN from the scratch, can be used to extract useful features for pain expression recognition in neonates. It also shows that combining deep features with traditional handcrafted features is a good practice to improve the performance of pain expression recognition, and possibly the performance of similar applications. 

As future work, we plan to fine tune the weights of the pre-trained CNNs by continuing the backpropagation. We also plan to incorporate other pain indicators (e.g., crying sound) into facial expression to develop a deep multimodal pain assessment system. 





\bibliographystyle{IEEEtran}
\bibliography{ref}

\end{document}